# Real-Time Structural Health Monitoring with Bayesian Neural Networks: Distinguishing Aleatoric and Epistemic Uncertainty for Digital Twin Frameworks

Hanbin Cho[a], Jecheon Yu[a], Hyeonbin Moon[a], Jiyoung Yoon[b], Junhyeong Lee[a], Giyoung Kim[c], Jinhyoung Park[d] and Seunghwa Ryu[a,e]∗

[a] Department of Mechanical Engineering, Korea Advanced Institute of Science and Technology (KAIST), Daejeon 34141, Republic of Korea

[b] Advanced Mechatronic R&D Group, Korea Institute of Industrial Technology, Daegu 42994, Republic of Korea

[c] Department of Mechanical Engineering, Kyungpook National University, Daegu 41566, Republic of Korea

[d] School of Mechatronics Engineering, Korea University of Technology and Education, Cheonan 31253, Republic of Korea

[e] KAIST InnoCORE PRISM-AI Center, Korea Advanced Institute of Science and Technology (KAIST), Daejeon, 34141, Republic of Korea

∗ Corresponding author. E-mail: ryush@kaist.ac.kr

# Abstract

Reliable real-time analysis of sensor data is essential for structural health monitoring (SHM) of high-value assets, yet a major challenge is to obtain spatially resolved full-field aleatoric and epistemic uncertainties for trustworthy decision-making. We present an integrated SHM framework that combines principal component analysis (PCA), a Bayesian neural network (BNN), and Hamiltonian Monte Carlo (HMC) inference, mapping sparse strain gauge measurements onto leading PCA modes to reconstruct full-field strain distributions with uncertainty quantification. The framework was validated through cyclic four-point bending tests on carbon fiber reinforced polymer (CFRP) specimens with varying crack lengths, achieving accurate strain field reconstruction ($R^2 > 0.9$) while simultaneously producing real-time uncertainty fields. A key contribution is that the BNN yields robust full-field strain reconstructions from noisy experimental data with crack-induced strain singularities, while also providing explicit representations of two complementary uncertainty fields. Considered jointly in full-field form, the aleatoric and epistemic uncertainty fields make it possible to diagnose at a local level, whether low-confidence regions are driven by data-inherent issues or by model-related limitations, thereby supporting reliable decision-making. Collectively, the results demonstrate that the proposed framework advances SHM toward trustworthy digital twin deployment and risk-aware structural diagnostics.

# 1. Introduction

Fatigue failure refers to the fracture of a structure or material under repeated loading. Unlike catastrophic failures caused by large, single loads, fatigue failure develops gradually under smaller loads applied over long durations, making it difficult to detect and prevent [1]. Conventional approaches generally divide into two categories: theoretical models based on microstructural observations or S-N curves [2,3], and data-driven lifetime prediction using machine learning [4-6]. Both, however, face limitations due to data scarcity, sensor noise, and model errors [7-9]. Rather than providing only one-off lifetime predictions, data-driven methods therefore need to be embedded in monitoring frameworks that track the evolving structural state over time. This perspective underscores the importance of structural health monitoring (SHM), in which the condition of a structure must be inferred efficiently from sparse sensor networks. For high-value assets, it is particularly critical to reconstruct full-field mechanical response in real time from limited strain measurements so that localized damage does not go undetected. At the same time, SHM frameworks must accompany each prediction with well-calibrated uncertainty explicitly indicating the risk of prediction failure, which is essential for credible decision-making in safety-critical settings. Meeting these requirements has motivated the development of digital twin (DT) frameworks, which have recently emerged as a promising solution for real-time monitoring and feedback of high-value assets.

A DT establishes a bidirectional link between a physical system and its virtual counterpart, enabling dynamic state estimation through continuous sensor-data integration [10-14]. Unlike conventional simulations, it operates as a dynamic system that continuously acquires sensor data to maintain accurate state estimation [15-20]. A fully functional DT framework requires the integration of three core technology domains: high-speed analysis techniques [8, 21-27], uncertainty quantification methods [28-33], and real-time decision-making algorithms [34,35]. Among these, real-time analysis and uncertainty quantification are critical prerequisites for trustworthy deployment in SHM [36,37] and constitute the primary focus of the present study.

Several studies have contributed to these enabling technologies. Pasparakis et al. [38] employed probabilistic estimation of composite material stress fields, applying methods such as Monte Carlo dropout and Bayes by Backprop to obtain full-field uncertainty estimates. Song et al. [39,40] applied dimension reduction-based Kriging to dynamical and time-varying systems, embedding latent features in Gaussian process kernels to enable efficient uncertainty propagation and sensitivity analysis. Zhang et al. [41] developed a finite element-based digital twin framework that reconstructs dynamic structural behavior from sparse displacement measurements, enabling estimation at non-sensor locations. Chen et al. [20] applied digital twin modeling to full-field reconstruction, pre-training a convolutional autoencoder on simulations and fusing sensor data via uncertainty-weighted losses to recover global stress and strain. Cao et al. [42] proposed a multiphase stochastic degradation framework that leverages similarity-based weighted likelihood estimation to extract both aleatory and epistemic uncertainties from degradation signals in real time.

To the best of our knowledge, no prior work in mechanical strain monitoring has presented an integrated framework that (i) reconstructs real-time full-field strain from sparse sensor data and (ii) simultaneously quantifies full-field aleatoric and epistemic uncertainties. Without full-field reconstruction capability, monitoring must rely only on a few aggregate signals, so localized damage or failure may remain undetected. In particular, obtaining and tracking full-fields of both uncertainties in real time makes predictions interpretable by localizing low-confidence regions and attributing them either to data-inherent noise (aleatoric) or to model insufficiency (epistemic), thereby supporting risk-aware decision-making.

To address this gap, we propose a real-time SHM framework that combines principal component analysis (PCA) with a Bayesian neural network (BNN) trained on sparse strain gauge measurements and refined through Hamiltonian Monte Carlo (HMC) posterior inference. The framework is validated using bending test data from carbon fiber reinforced polymer (CFRP) specimens with various crack lengths, as detailed in the subsequent section. A key novelty of the framework is a mode-aware probabilistic formulation that explicitly accounts for the unequal importance of the PCA modes. During BNN training, a separate variance is learned for each mode, and these mode-wise variances are then used as likelihood scales in HMC. This design directly anchors the sampling of epistemic uncertainty in the aleatoric variability of the data. This approach is particularly suitable for systems with outputs of differing scale and importance. It optimizes a separate variance for each mode and incorporates these mode-wise variances into the HMC likelihood, leading to reliable and data-consistent posterior sampling. **Fig. 1** provides a schematic overview of the overall framework. PCA supplies a compact basis for strain field representation and is particularly suitable for time-continuous strain fields. Because the experimental dataset is noisy and exhibits crack-induced strain singularities, a BNN that robustly maps sparse sensor data to the leading PCA modes is employed to handle these effects. Although HMC entails higher computational cost than alternative inference methods, its superior estimation accuracy motivates its adoption in the present framework, which targets high-value assets. Together, these components form a probabilistic formulation that supports both accurate full-field reconstruction and uncertainty-aware diagnostics, paving the way toward reliable digital twin systems for SHM.

## 2. Experimental data

The experimental dataset used in this study was adopted from Yoon et al. [43], where cyclic four-point bending tests were performed on CFRP specimens with and without edge cracks. CFRPs are widely used as primary load-bearing materials in high-performance structures, including aircraft fuselage and wings, wind-turbine blades, and retrofitted civil infrastructure, owing to their high specific stiffness, strength, and corrosion resistance [44-46]. Cyclic four-point bending of CFRP specimens provides a representative loading scenario for practical fatigue problems in composite structures. In particular, Yoon et al. [43] regarded this configuration as a laboratory analogue of the bending state experienced by an aircraft wing under gravity and aerodynamic lift. Eight specimens were tested in total: one healthy specimen and seven damaged specimens with crack lengths ranging from 1 to 15 mm. Detailed specimen specifications are provided in **Supplementary A**. Mechanical loading was applied using a uniaxial hydraulic testing machine (MTS Systems) under cyclic four-point bending between 0 and 20 mm at 0.1 Hz, representing quasi-static fatigue conditions.

Strain measurements were acquired using two complementary techniques. Local strain was recorded from 12 uniaxial strain gauge sensors distributed across the specimen surface, while full-field strain was captured using a digital image correlation (DIC) system with a 14 × 12 subset grid, yielding 168 spatial subsets. The overall measurement setup, including specimen geometry, strain gauge locations, and DIC coverage, is shown in **Fig. 1 (a)**.

For each specimen type, five independent tests were conducted, resulting in approximately 40,000 synchronized data samples. Each sample consisted of paired strain gauge readings and the corresponding DIC full-field strain field. This dataset captures both global deformation and local crack-tip strain under cyclic loading and serves as the basis for training and evaluating the proposed SHM framework. For each specimen type, data from four experiments were used for model training, whereas the remaining one experiment was reserved for independent testing. Building on this dataset, we next describe the methodology for dimensionality reduction, probabilistic modeling, and uncertainty quantification.

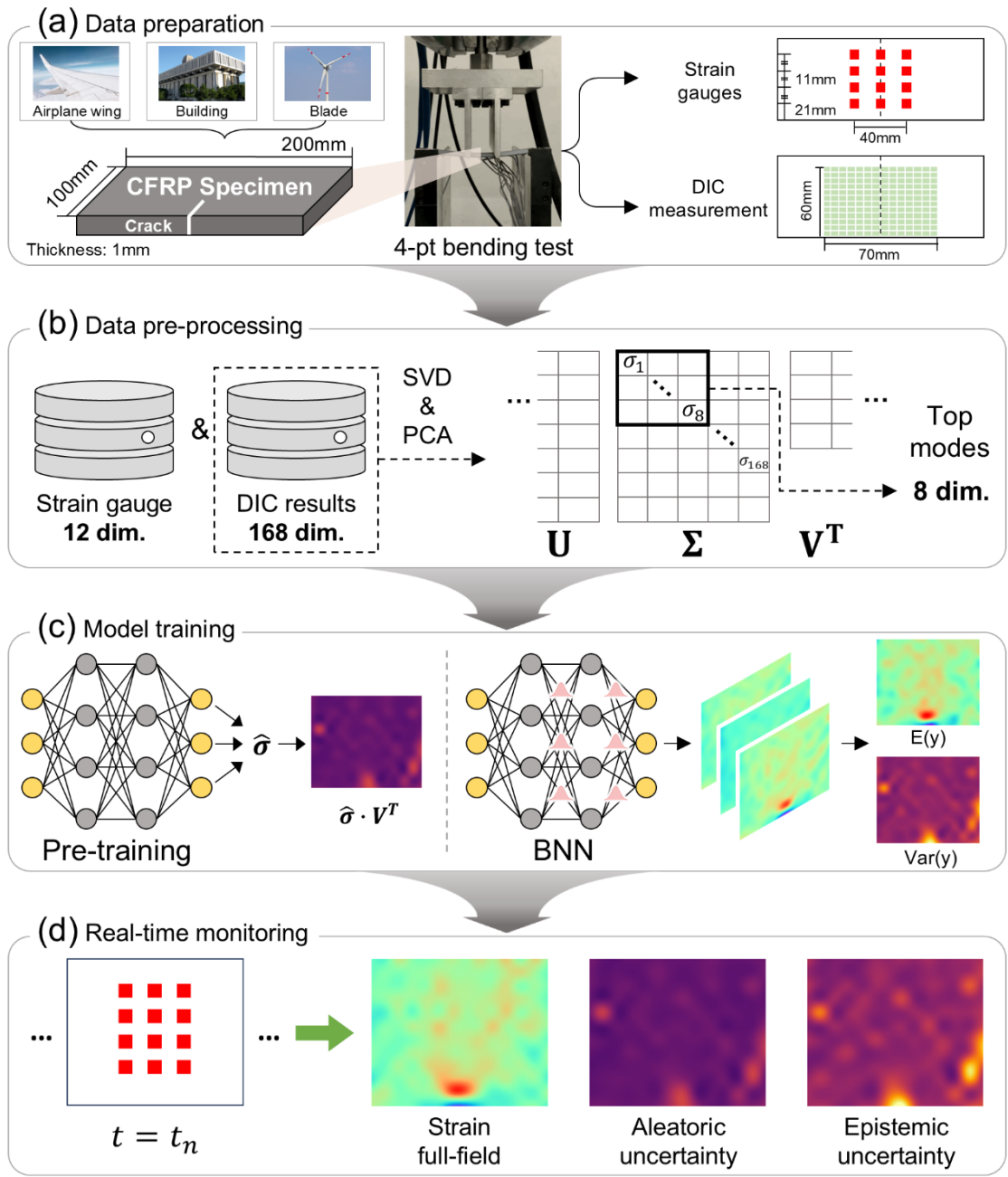

**Figure 1** Schematic of the proposed real-time SHM framework with uncertainty quantification. (a) Data preparation: acquisition under four-point bending test using two measurement modalities, 12 strain gauges and DIC full-field strain on a 14 × 12 grid, 168 subsets. (b) Data preprocessing for dimension reduction: the 168-dimensional DIC full-field strain is reduced to eight dimensions using PCA to enhance predictive performance. (c) BNN training: parameter distributions are inferred, enabling simultaneous prediction and uncertainty quantification

through mean and variance. (d) Real-time monitoring: data from sparse strain gauges are processed by the trained BNN to simultaneously reconstruct the full-field strain and quantify aleatoric and epistemic uncertainty fields.

## 3. Method

### 3.1. Principal component analysis

The full-field strain data obtained from DIC were represented as 168-dimensional vectors, corresponding to the 14 × 12 grid of spatial subsets. Directly training a surrogate model to predict such high-dimensional outputs from sparse sensor inputs will typically result in poor accuracy and efficiency due to the curse of dimensionality [47]. To address this, PCA was employed to reduce the output dimensionality while preserving the essential structure of the strain field, as shown in **Fig. 1 (b)**.

The DIC dataset, composed of $N$ samples, was first min-max normalized across all subsets. The resulting data matrix $\mathbf{X} \in \mathbb{R}^{N \times 168}$ was then decomposed via singular value decomposition (SVD):

$$\mathbf{X} = \mathbf{U}\mathbf{\Sigma}\mathbf{V}^\mathrm{T} \tag{1}$$

Here, $\mathbf{U} \in \mathbb{R}^{N \times N}$ and $\mathbf{V} \in \mathbb{R}^{168 \times 168}$ are orthogonal matrices containing the left and right singular vectors, respectively, and $\mathbf{\Sigma}$ is a diagonal matrix whose elements are the singular values, representing the square roots of the eigenvalues of the covariance matrix.

Dimensionality reduction was performed by selecting the top $k$ principal components, projecting the original strain fields into a lower-dimensional latent space. This reduced representation captures the dominant modes of variation across the dataset. The choice of $k$ balances reconstruction fidelity and model tractability: larger $k$ improves accuracy but increases complexity, while smaller $k$ reduces complexity at the cost of detail. To balance this trade-off, the cumulative explained variance (CEV) ratio was computed. Formally, letting $\{\lambda_i\}_{i=1}^{p}$ denote the eigenvalues of the covariance matrix in descending order, the CEV is as follows:

$$\text{CEV}(k) = \frac{\sum_{i=1}^{k} \lambda_i}{\sum_{j=1}^{p} \lambda_j} = \frac{\sum_{i=1}^{k} \sigma_i^2}{\sum_{j=1}^{p} \sigma_j^2} \qquad (2)$$

where $\sigma_i$ denotes the square root of the $i$-th eigenvalue of the covariance matrix $\Sigma$ under SVD, and $p$ refers to the total dimension of the target space. As shown in **Fig. 2 (a)**, the top eight principal components were selected, which collectively account for approximately 95% of the total variance in the original DIC data. This compact representation allows the surrogate model to operate in an eight-dimensional output space, from which the 168-dimensional strain field can be accurately reconstructed with minimal loss of structural detail. **Fig. 2 (b)** shows the true strain curve along with the restored curves for different numbers of modes. While slight fluctuations remain, the eight leading modes were sufficient to accurately restore the actual strain curve.

The PCA coefficients for each sample are given by $\boldsymbol{Z} = \boldsymbol{U}\boldsymbol{\Sigma}$, and the reduced PCA coefficients are $\boldsymbol{Z}_r = \boldsymbol{U}_r \boldsymbol{\Sigma}_r \in \mathbb{R}^{N \times k}$, corresponding to the top $k$ principal components. To learn the mapping from sparse strain gauge inputs to reduced PCA coefficients, BNN was employed. The BNN took a 12 dimensional strain gauge vector as input and predicted eight PCA coefficients, while its trainable parameters (weights and biases across hidden layers) were treated as random variables. Since reliable uncertainty quantification requires posterior inference over these parameters, HMC was adopted as the sampling method, as detailed in the following sections.

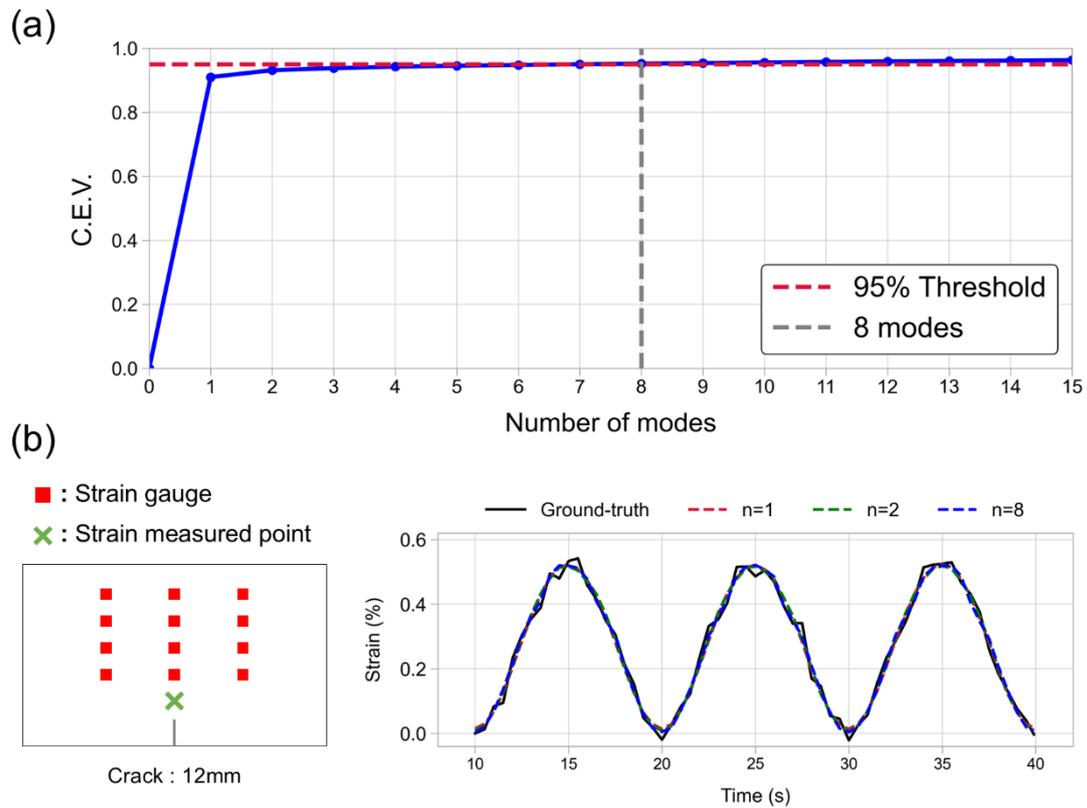

**Figure 2** Selection of the number of PCA modes. (a) Cumulative explained variance as a function of the number of principal components. The dashed line indicates the 95% cumulative variance threshold, and the green dashed line marks the smallest number of modes that satisfies the criterion, eight modes. (b) To illustrate the reconstruction accuracy with different numbers of modes, restored strain curves at the location marked with a green "×", obtained using different numbers of modes.

### 3.2. Bayesian neural network

The BNN served as the core predictive model, mapping 12 dimensional strain gauge inputs to 8 PCA coefficients that represent the reduced strain field. Unlike conventional deterministic neural networks, the BNN adopts a probabilistic formulation in which its parameters (weights and biases) are treated as random variables. The posterior distribution over

the parameters $\boldsymbol{\theta}$ given the dataset $\mathcal{D}$ is defined via Bayes' theorem:

$$p(\boldsymbol{\theta}|\mathcal{D}) = \frac{p(\mathcal{D}|\boldsymbol{\theta})p(\boldsymbol{\theta})}{p(\mathcal{D})} \qquad (3)$$

where $p(\boldsymbol{\theta}|\mathcal{D})$ denotes the posterior distribution, $p(\mathcal{D}|\boldsymbol{\theta})$ is the likelihood function, and $p(\boldsymbol{\theta})$ represents the prior distribution over the model parameters. The term $p(\mathcal{D})$ is the marginal probability. This probabilistic formulation enables simultaneous, full-field quantification of aleatoric uncertainty learned during pre-training and epistemic uncertainty obtained via HMC posterior inference. Compared to conventional Markov Chain Monte Carlo (MCMC) or Metropolis-Hastings algorithms, HMC achieves faster convergence and better sampling efficiency in high-dimensional parameter spaces [48, 49]. Variational inference (VI) approximates the posterior within a restricted family, trading some accuracy for lower computational cost and improved convergence. However, for high-value asset SHM where predictive accuracy and well-calibrated uncertainty are paramount, HMC remains the safer and more appropriate choice despite its higher computational burden. The theoretical background of HMC is detailed in **Supplementary B**.

The network architecture comprised three fully connected hidden layers with 100 neurons each, and posterior inference was performed using HMC. From the four experiments designated for model training (as described in **Section 2**), the dataset was further randomly divided into training and validation subsets with a ratio of 8:2 for the pre-training stage. To stabilize this process, a pre-training was conducted using the Adam optimizer with a learning rate of 0.001 over 300 training epochs.

To minimize prediction error while enabling uncertainty quantification, a modified loss function was employed that incorporates both the mode-wise importance of PCA components [43] and a probabilistic treatment of aleatoric uncertainty [50]. A Gaussian likelihood is assumed for each PCA coefficient, and the mean together with a mode-wise variance are learned, thereby assigning a data-driven variance to every output. The resulting mode-wise predictive variances approximate data-inherent, heteroscedastic aleatoric uncertainty. The loss function was defined as follows:

$$\mathcal{L} = \frac{1}{N}\frac{1}{8}\sum_{i=1}^{N}\sum_{j=1}^{8}\frac{\lambda_j}{\sum_{k=1}^{168}\lambda_k}\left[\frac{1}{2}\hat{\sigma}_j^{-2}(z_i[j]-\hat{z}_i[j])^2+\frac{1}{2}\log\hat{\sigma}_j^2\right] \quad (4)$$

where $z_i[j]$ and $\hat{z}_i[j]$ denote the *j*-th PCA coefficient of the *i*-th true and predicted data, respectively. The *N* refers to the total number of data. The term $\lambda_k$ represents the squared value of the *k*-th PCA mode, reflecting its explained variance. Consequently, the weighting term $\frac{\lambda_j}{\sum \lambda_k}$ normalizes the explained variance of each mode, allowing the loss function to emphasize the dominant modes during training.

The first term inside the summation corresponds to the negative log-likelihood of a Gaussian distribution, enabling aleatoric uncertainty to be learned by associating a predicted variance $\hat{\sigma}_j^2$ with each mode. The second term, $\frac{1}{2}\log\hat{\sigma}_j^2$, acts as a regularization penalty to prevent the model from inflating variance estimates arbitrarily. This formulation ensures that the network not only minimizes data error but also properly calibrates the inherent uncertainty associated with each mode. While this pre-training step may limit exploration of the full posterior, it is essential for ensuring efficient and stable sampling in high-dimensional parameter spaces [38]. The resulting parameters served both as the initial state for HMC sampling and the mean of the prior distribution, enabling more reliable posterior estimation.

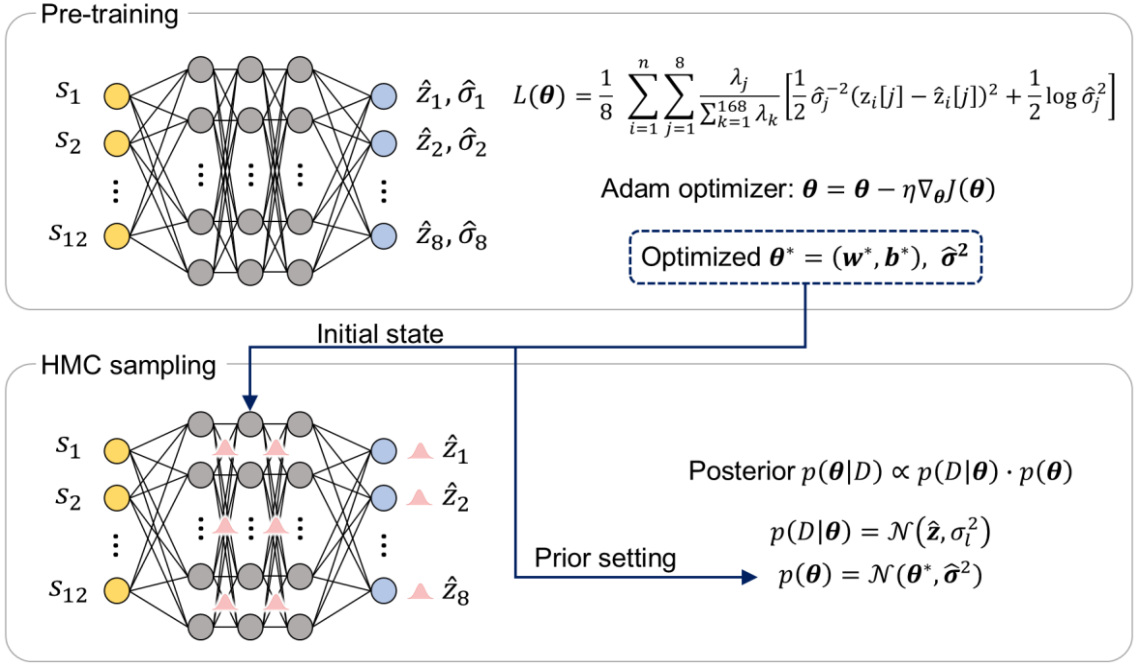

**Figure 3** Schematic of the BNN training process. The BNN is trained in two stages. In the first stage, pre-training with the Adam optimizer minimizes error while learning mode-wise variances, producing a stable parameter initialization and quantifying aleatoric uncertainty. In the second stage, HMC sampling infers the posterior distribution of all trainable parameters, enabling calibrated estimation of epistemic uncertainty. The figure illustrates how the pre-trained weights serve both as initialization and as prior means for posterior sampling.

Next, parameter estimation was performed using HMC sampling to draw samples from the posterior over all trainable network parameters, including weights and biases. Independent Gaussian priors centered at the pre-trained estimate $\boldsymbol{\theta}^*$ with a constant standard deviation were placed on all parameters. In this study, a value of 0.5 was selected as the constant standard deviation for the prior distribution, which is smaller than the unit Gaussian commonly used in previous studies [31,51]. The value was adopted since pre-training had already provided a well-initialized parameter space. Then, HMC sampling was initialized at $\boldsymbol{\theta}^*$ to promote stable and efficient convergence. The overall negative log-likelihood was expressed as the product of the weighted, mode-wise likelihoods:

$$-\log \mathcal{L}(\boldsymbol{\theta}) = \sum_{i=1}^{8} \left( N \log \hat{\beta}_i + \frac{1}{2\hat{\beta}_i^2} \sum_{j=1}^{N} \left( \mu_{ij} - f_i(\boldsymbol{s}_j; \boldsymbol{\theta}) \right)^2 \right) \quad (5)$$

Here, $\mu_{ij}$ is the ground-truth value of the $j$-th training sample in the $i$-th mode, $\boldsymbol{s}_j$ is the corresponding strain gauge input, and $f_i(\boldsymbol{s}_j; \boldsymbol{\theta})$ is the BNN prediction for $i$-th mode given parameters $\boldsymbol{\theta}$. The Gaussian likelihood used a mode-wise standard deviation

$$\beta_i = \hat{\sigma}_i / d \quad (6)$$

which means the likelihood scale for $i$-th mode is the predicted standard deviation $\hat{\sigma}_i$, resulted from Eq. (4) divided by a positive constant $d$. Since the variance for each output variable was calculated during pre-training, these values can be directly used as the likelihood standard deviations. However, while the mode-wise variances account for the importance of each variable, they do not account for the scale difference with the prior distribution variance. To address this, a global calibration factor $d$ was introduced. In this study, a value of 20 was selected as $d$, which balances the likelihood variance and prior variance during HMC sampling.

The BNN training process is visualized in **Fig. 3**, which includes both the pre-training process and HMC sampling, with the initial state being the result of the pre-training. The sampler was configured with a burn-in period of 100 iterations, after which 1,000 samples were collected for posterior inference. Since the initial state and the prior mean can be assumed to be reasonable, the number of burn-in iterations was minimized to improve sampling efficiency. The step size and target acceptance ratio were initially set to 0.0001 and 0.6, respectively, to promote stable and efficient sampling.

The overall workflow of the proposed SHM framework is summarized in **Fig. 4**. The high-dimensional strain fields from DIC were reduced using PCA, the BNN mapped the sparse strain gauge inputs to these reduced PCA coefficients, and HMC sampling was employed for posterior inference. Once trained, the model reconstructed a full-field strain distributions in real-time while providing spatially resolved aleatoric and epistemic uncertainty fields. This combined capability enabled accurate reconstruction, interpretable diagnostics, and reliable integration into digital twin systems.

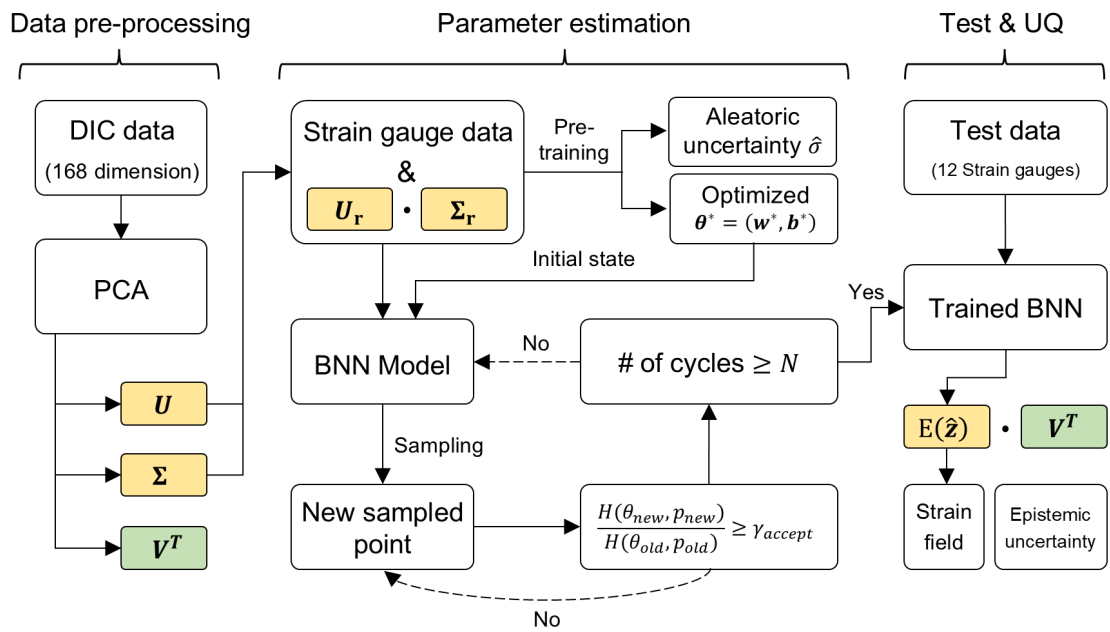

**Figure 4** Workflow of the proposed SHM framework integrating PCA, BNN, and HMC sampling. The process begins with dimensionality reduction of DIC full-field strain via PCA, providing a compact basis for surrogate modeling. Strain gauge data are then mapped to the reduced PCA coefficients using a BNN, with pre-training to learn mode-wise variance for aleatoric uncertainty and HMC sampling for posterior inference of network parameters. The trained model reconstructs full-field strain distributions from sparse strain gauge inputs while producing spatially resolved aleatoric and epistemic uncertainty estimates, enabling real-time uncertainty-aware monitoring.

# 4. Results

## 4.1. *Aleatoric* uncertainty learned during pre-training

Aleatoric uncertainty, arising from inherent measurement noise and sensor configuration [52], was quantified during the BNN pre-training stage. **Table 1** summarizes the mode-wise predicted variances $\hat{\sigma}_j$ obtained from Eq. (4). Consistent with the explained variance ratios shown in **Fig. 2 (a)**, the leading modes, which contribute most to strain field reconstruction, exhibited relatively smaller learned variances, reflecting higher model confidence. In contrast, higher order modes showed larger variances, consistent with their limited contribution to overall reconstruction fidelity. Each mode-wise predicted variance represented the level of inherent uncertainty associated with its corresponding principal component. The training and validation loss curves from the pre-training process are shown in **Supplementary C**.

**Table 1**. The eight mode-wise predicted variance values resulted from the pre-training phase.

| Index | 1 | 2 | 3 | 4 | 5 | 6 | 7 | 8 |
|---|---|---|---|---|---|---|---|---|
| **Value** | 0.024 | 0.019 | 0.036 | 0.038 | 0.034 | 0.031 | 0.043 | 0.058 |

Then, the aleatoric uncertainty field was reconstructed through the following expression:

$$\hat{\boldsymbol{\sigma}} \cdot \boldsymbol{V}^T \tag{7}$$

The resulting full-field aleatoric uncertainty is shown in **Fig. 5**. The 168-dimensional data mapped to the 14 × 12 grid were subsequently upsampled and smoothed via bicubic interpolation [53] to suppress high-frequency noise and enhance the visibility of crack-tip strain concentrations, thereby facilitating more reliable localization of critical regions. Two regions exhibited distinctly elevated uncertainty: (i) near the crack tip, where sharp strain gradients and specimen-to-specimen crack length variability induced high heteroscedastic variance, and (ii) along the upper-left edge, the region far from the strain gauges, where weak input-output correlation necessitated extrapolation. Additionally, as shown in **Fig. 6**, a subset of the DIC training data exhibited low strain near the upper-left edge. This region corresponded to elevated aleatoric uncertainty. Moreover, as illustrated in **Fig. 6 (c)**, values close to outliers were occasionally observed in that region. Because aleatoric uncertainty is learned during pre-training and reflects such data-inherent characteristics, it served as an early indicator of inherent data abnormalities. Taken together, these observations indicated that it captured regions dominated by noise structure, sensor sparsity, and basis truncation effects and provided a physics-consistent interpretation of where deterministic predictions were less reliable.

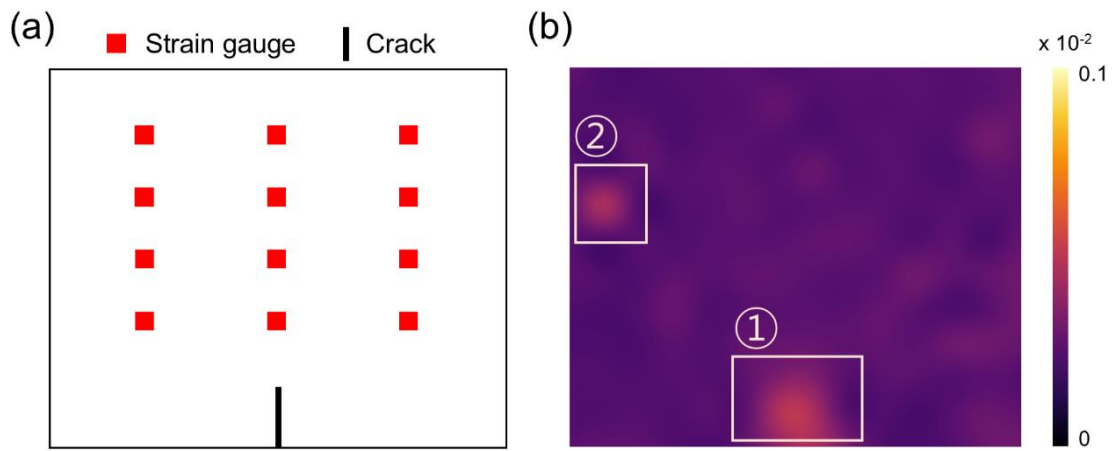

**Figure 5** Aleatoric uncertainty field with the corresponding strain gauge layout for location-

specific interpretation. (a) Strain gauge arrangement and crack location within the DIC target area of the specimen. (b) Aleatoric uncertainty field derived from mode-wise predicted variances, highlighting two regions with particularly high uncertainty.

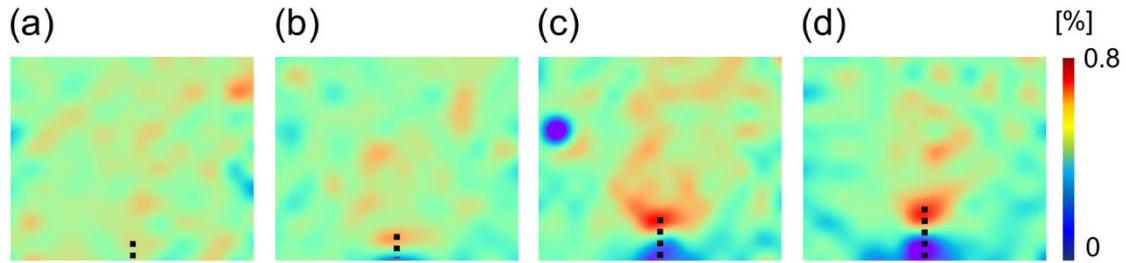

**Figure 6** Ground-truth strain fields obtained via DIC and used for training under maximum-displacement loading, showing abnormal strain values near the upper-left edge. Black dashed lines represent the crack location and length for each specimen. (a) Specimen with 5mm crack; (b) Specimen with 7mm crack; (c) Specimen with 12mm crack; (d) Specimen with 15mm crack.

In contrast to previous HMC-based studies that assigned a single likelihood variance in a heuristic manner [49, 54-55], the present work employed a mode-wise variance vector guided by the PCA output structure. This formulation reflected the unequal importance of individual modes and anchored the likelihood specification in the aleatoric variability learned during pre-training. Consequently, the proposed approach replaced a manually chosen single-variance settings with mode-aware, data-driven likelihood scales, ensuring consistency between the statistical model and the underlying physics.

**4.2. Strain field reconstruction and *epistemic* uncertainty**

Posterior inference through HMC sampling enabled real-time reconstruction of full-field strain together with calibrated estimates of epistemic uncertainty. On a workstation equipped with a 13th Gen Intel Core i9-13900K CPU, end-to-end inference on the test dataset (from input ingestion to high-resolution result visualization) averaged 0.53 s per sample, indicating practical feasibility for real-time analysis. Implementation details and the timing breakdown are provided in **Supplementary D**. **Fig. 7** and **8** presents results for the test experiments partitioned in **Section 2**, comparing the reconstructed strain fields with ground-truth DIC measurements for specimens of varying crack lengths under maximum displacement loading. As in the previous section, the DIC data collected on a 14 × 12 grid was upsampled and smoothed using bicubic interpolation. The BNN-predicted mean strain fields are shown in **Fig. 7 (a)** and **Fig. 8 (a)**, while the corresponding ground-truth DIC fields are presented in **Fig. 7 (b)** and **Fig. 8 (b)**, respectively. The strain field reconstruction achieved by the proposed framework captured the overall distribution of the strain field across the entire target area. Despite local discrepancies in boundary regions or near cracks, the reconstructed fields were shown to match the ground-truth DIC measurements with sufficient fidelity for structural health monitoring purposes. Based on a qualitative assessment of the predicted mean field in the crack region shown in **Fig. 7 (a)** and **Fig. 8 (a)**, the framework appeared to distinguish different crack lengths by capturing characteristic variations in the strain field, thus validating its capability to reproduce a key observation reported in prior work [43].

The absolute error fields shown in **Fig. 7 (c)** and **Fig. 8 (c)** indicated that discrepancies were largest near specimen boundaries and crack tips, where reconstruction was inherently difficult due to strain singularities and optical distortion effects in DIC. In addition, large error regions tended to appear near the crack as the crack length increased. These errors arose from the strain field singularities and discontinuities introduced by the crack, which hindered accurate reconstruction of the strain field from strain gauge in such regions.

The epistemic uncertainty fields shown in **Fig. 7 (d)** and **Fig. 8 (d)** were derived from posterior sample variability and reconstructed via the following expression:

$$\text{Std}(\hat{\boldsymbol{z}}) \cdot \boldsymbol{V}^T \qquad (8)$$

where $\text{Std}(\hat{\boldsymbol{z}})$ denotes the standard deviation of the predicted PCA coefficient vector across posterior samples. Regions near crack tips and at the periphery of the sensing domain exhibited elevated epistemic uncertainty, indicating reduced model confidence. However, no clear correlation between crack length and the overall magnitude of epistemic uncertainty was observed; differences among the fields in **Fig. 7 (d)** and **Fig. 8 (d)** appeared to be driven primarily by noise characteristics in the training data and by the specific test inputs, rather than by crack length itself. These results align with the definition of epistemic uncertainty, which represents model confidence and reflects the input-dependent risk of prediction failure. Furthermore, **Supplementary Video 1** visualizes the temporal progression of the BNN-predicted strain field for a specimen with a 12mm crack, together with the associated uncertainty and error fields, and the ground-truth strain field for comparison. Details on validating the BNN results by comparison with another prominent uncertainty quantification method, Monte Carlo dropout, are provided in **Supplementary E**.

To examine the field-level errors in **Fig. 7 (c)** and **Fig. 8 (c)** more closely, **Fig. 9** shows the BNN predictions for the PCA coefficients under maximum displacement loading. Across specimens, most cases exhibited low MAE and RMSE, and R square values exceeded 0.9 for all evaluations. Thus, the errors observed in **Fig. 7 (c)** and **Fig. 8 (c)** reflected both BNN prediction errors and errors due to information loss incurred during reconstruction from the truncated PCA basis. The complete sequence of PCA coefficient predictions for the 12mm crack specimen over the entire bending test is provided in **Supplementary Video 2**, with its corresponding true strain fields, to clarify the loading steps. As crack length increased, the symmetry of the overall strain field was emphasized due to the singularity at the crack tip, although noisy and localized deformations also became more prominent. As shown in **Fig. 9**, the accuracy of the top PCA coefficients, which have a significant influence on the reconstruction, remains high across all specimens. However, as crack length increases, the fluctuations in the lower PCA coefficients, which capture localized variations, become more pronounced, leading to a slight decrease in reconstruction accuracy.

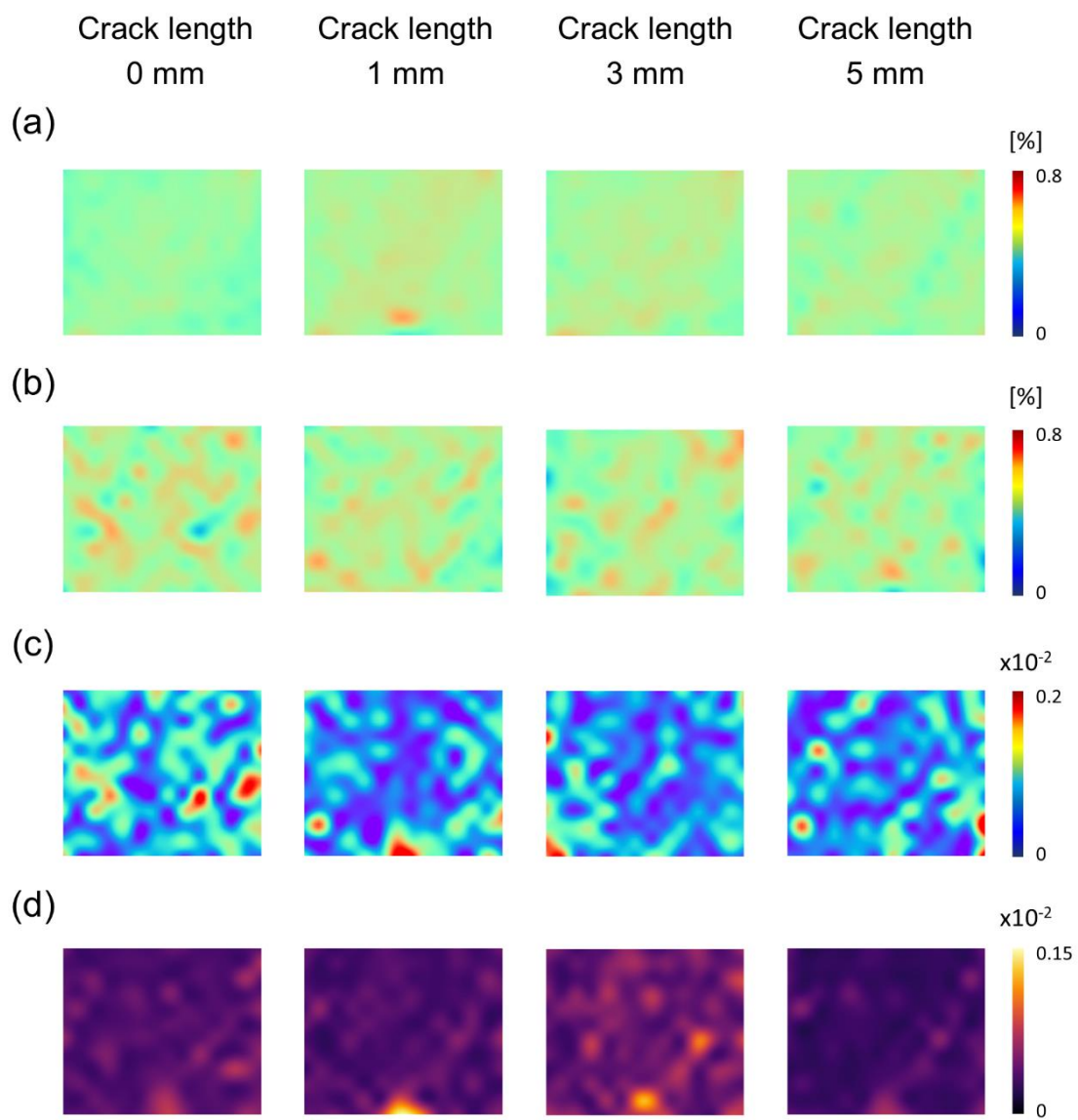

**Figure 7** The restored strain field results for the four smaller crack lengths among the eight

cases. (a) The strain field, predicted and restored using BNN based on the PCA coefficients, expressed as the mean value. (b) The true strain field measured via DIC. (c) The absolute error field, showing the difference between the predicted/restored strain field and the true strain field. (d) The epistemic uncertainty field corresponding to the predicted/restored strain field.

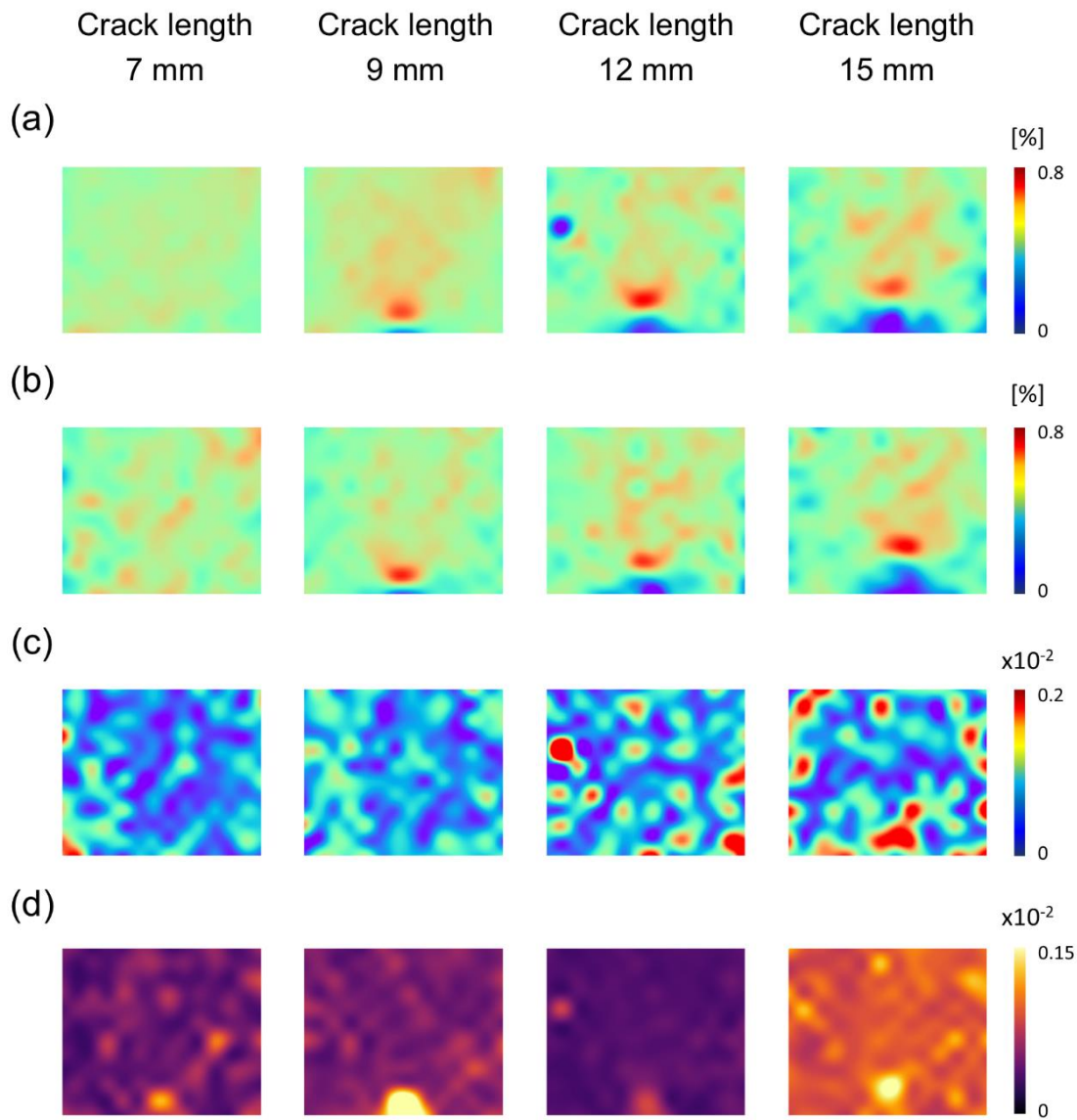

**Figure 8** The restored strain field results for the four larger crack lengths among the eight cases. (a) The strain field, predicted and restored using BNN based on the PCA coefficients, expressed as the mean value. (b) The true strain field measured via DIC. (c) The absolute error field,

showing the difference between the predicted/restored strain field and the true strain field. (d) The epistemic uncertainty field corresponding to the predicted/restored strain field.

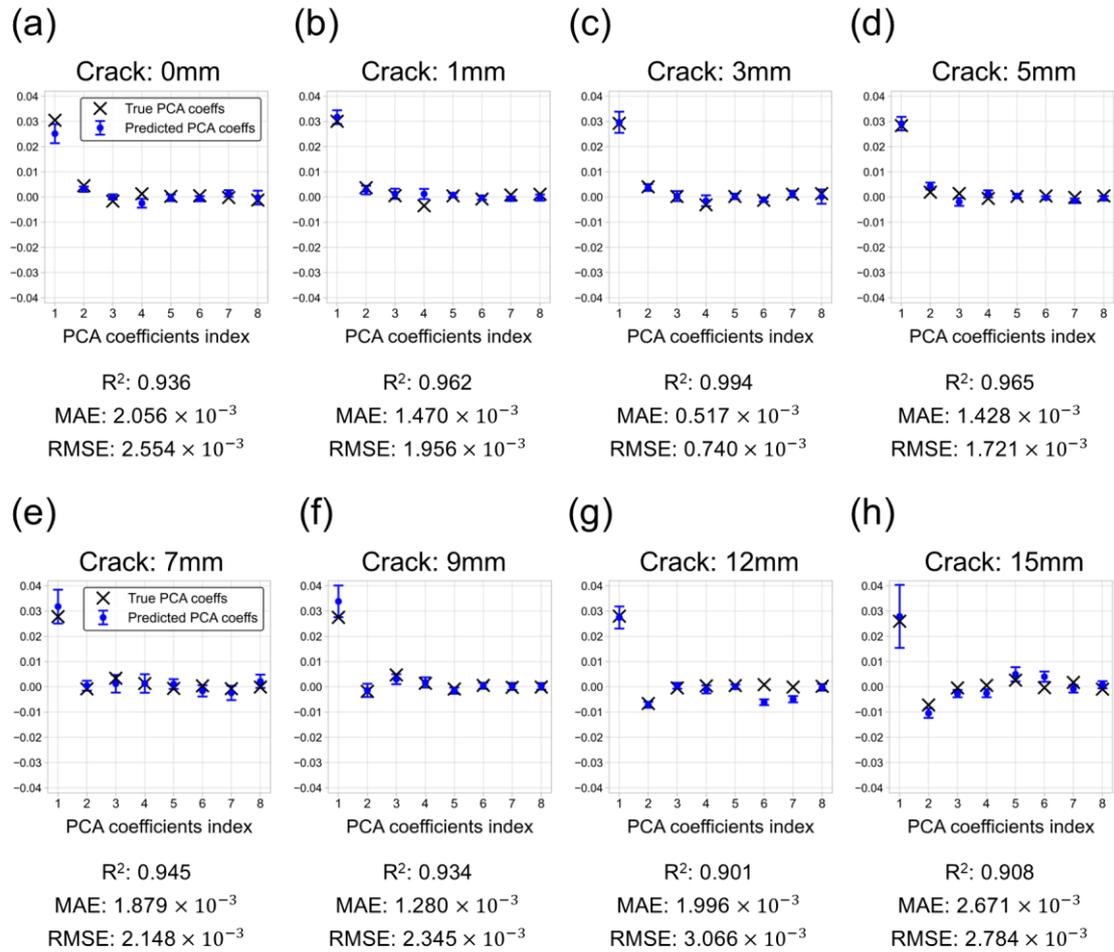

**Figure 9** Prediction of PCA coefficients under maximum-displacement loading for each specimen, shown with BNN-derived 95% confidence intervals and three error metrics. (a–h) Results for specimens at each crack length.

### 4.3. Analysis of uncertainty characteristics and complementarity

Comparing the aleatoric uncertainty field in **Fig. 5** with the epistemic uncertainty fields in **Fig. 7 (d)** and **Fig. 8 (d)**, both exhibited elevated values near the crack tip. However, whereas the aleatoric field learned during pre-training remained fixed across test datasets, the epistemic field varied with the specific test input. When measurement noise was likely or the strain field exhibited large spatial or temporal changes, the epistemic uncertainty showed a larger overall magnitude. This input dependence was further illustrated in **Supplementary Video 1**, which showed that under maximum-displacement loading the strain field activated across the entire target area and the epistemic uncertainty field likewise increased; upon unloading, as the strain field returned to near-zero levels, the epistemic uncertainty field diminished to nearly zero throughout the domain.

In **Fig. 8 (a)**, for the 12 mm crack specimen, an anomalously low region appeared near the upper-left edge of the BNN-predicted mean strain field. While high aleatoric uncertainty near sensor-distant boundaries reflected low confidence, the elevated predicted mean strain in this region also suggested outliers in the measurement data. For the 12mm specimen, abnormal training samples were present around that area, so the aleatoric uncertainty was high, and the discrepancy between abnormal and nominal samples caused the epistemic uncertainty to increase as well, as shown in **Fig. 8 (d)**. Consequently, this behavior stemmed from defects inherent in the data rather than insufficient model capability. These defects, in turn, caused issues in specific parts of the model training process. Within a real-time SHM framework, the aleatoric and epistemic uncertainty fields play complementary roles. Because the aleatoric uncertainty field remains fixed across test inputs, a region exhibiting high aleatoric uncertainty together with low epistemic uncertainty indicates data-inherent issues (e.g., sensor drift, sporadic noise, or outliers); in practice, this prompts actions such as sensor quality checks or outlier removal to reduce the risk of prediction failure. Conversely, regions with elevated epistemic uncertainty indicate model insufficiency under the current operating conditions, motivating targeted data acquisition around similar states and subsequent model updates.

Accordingly, obtaining both fields is essential for effective and efficient real-time SHM, and their joint use naturally interfaces with the decision layer of a DT pipeline to enable risk-aware monitoring and maintenance planning.

Within this analysis, uncertainty quantification complemented predictions by enabling continuous tracking of structural state in the presence of sensor noise and different crack length of each specimen. Consequently, the principal value of the uncertainty field lay less in its absolute magnitude than in its capacity to track state changes over time. The utility of such fields was further amplified when integrated into a monitoring framework such as a digital twin, enabling risk-aware diagnostics and decision-making. The method for merging the aleatoric and epistemic uncertainty fields into a single unified uncertainty field, together with illustrative examples, is provided in **Supplementary F**.

**4.4 Influence of training data size on *aleatoric* and *epistemic* uncertainties**

The variation of both aleatoric and epistemic uncertainty fields with respect to the number of training data was examined. First, note that aleatoric uncertainty, by definition, is independent of the number of data samples. The number of training samples was increased from 300 to 900 in increments of 200. Since the aleatoric uncertainty field is derived during pre-training, pre-training was conducted separately for each dataset, and the mode-wise predicted variances obtained in each case were utilized.

**Table 2** The eight mode-wise predicted variance values obtained from the pre-training phase for each training data size.

| Index | 1 | 2 | 3 | 4 | 5 | 6 | 7 | 8 |
|---|---|---|---|---|---|---|---|---|
| **300** | 0.025 | 0.024 | 0.047 | 0.049 | 0.041 | 0.038 | 0.050 | 0.072 |
| **500** | 0.024 | 0.019 | 0.036 | 0.038 | 0.034 | 0.031 | 0.043 | 0.058 |
| **700** | 0.023 | 0.019 | 0.034 | 0.035 | 0.035 | 0.031 | 0.040 | 0.053 |
| **900** | 0.023 | 0.018 | 0.032 | 0.033 | 0.032 | 0.030 | 0.038 | 0.052 |

**Table 2** lists the eight mode-wise variances obtained from the pre-training process for each training sample size. As shown in the **Table 2**, the eight mode-wise variance values changed only marginally with training sample size, and the variation became negligible once more than 500 samples per experiment are used. This behavior indicated that, when the dataset is too small, aleatoric uncertainty cannot be quantified reliably. However, once the data size exceeds a sufficient threshold, the learned variances become effectively independent of sample size. In this sense, the aleatoric uncertainty obtained with the proposed approach was consistent with its theoretical definition.

To examine the aleatoric uncertainty field corresponding to each set of mode-wise variances, these variances were reconstructed to the full-field uncertainty map. As shown in **Fig. 10**, the results indicated that the overall magnitude of the aleatoric uncertainty field decreased as the number of training samples increased. To present this trend more clearly, **Fig. 11** shows histograms and median values for the four cases. As illustrated in **Fig. 11**, both the overall magnitude of the aleatoric uncertainty field and its median value decreased with increasing training sample size.

As mentioned above, aleatoric uncertainty is inherent to the data generating process and should not depend on the number of samples. In the present implementation, the learned mode-wise variances were effectively independent of training-sample size, whereas the corresponding reconstructed full-field aleatoric uncertainty exhibited a mild dependence due to changes in the reconstruction basis. This arose because the PCA basis and the associated reconstruction matrix, which were used to obtain the full-field aleatoric uncertainty from the mode-wise variances, were re-estimated for each dataset and thus changed slightly as more data became available. Importantly, however, the spatial pattern of the aleatoric uncertainty field remained essentially unchanged and only its overall magnitude shifted. Therefore, the interpretation should focus on the spatial distribution of high-uncertainty regions rather than the absolute magnitude of the aleatoric field.

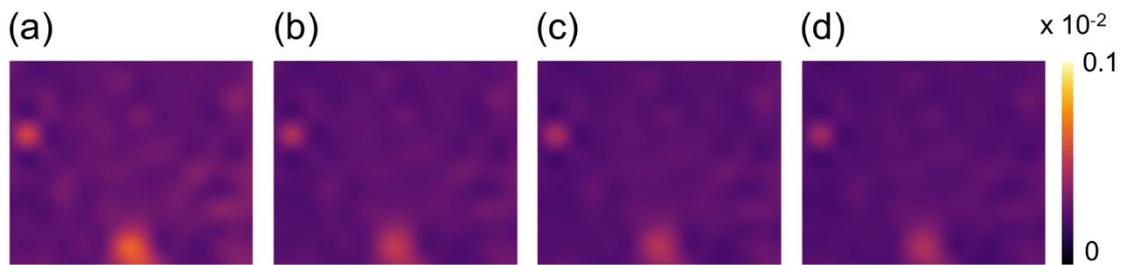

**Figure 10** Aleatoric uncertainty fields for different numbers of training samples per experiment. (a) 300; (b) 500; (c) 700; (d) 900.

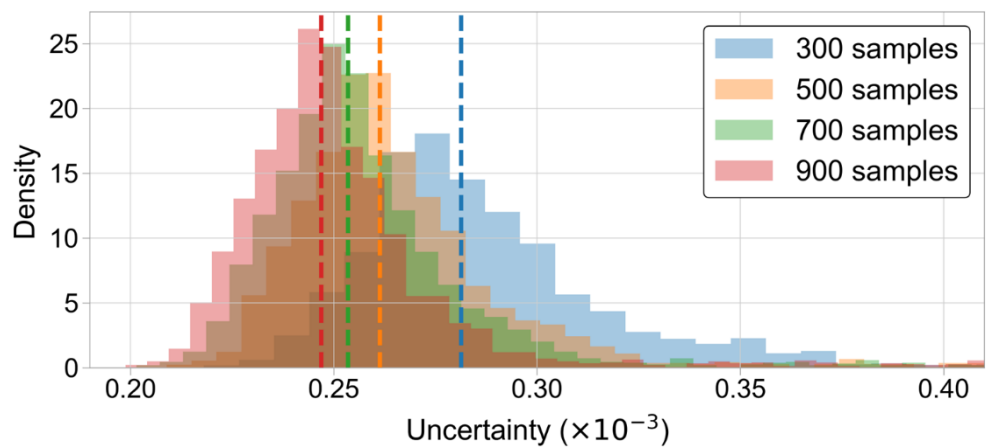

**Figure 11** Histograms and median values of aleatoric uncertainty fields for four cases, each based on varying amounts of training data used per experiment.

On the other hand, epistemic uncertainty is modulated by data availability, unlike aleatoric uncertainty. Varying the number of training samples per experiment revealed a clear data dependence in the epistemic uncertainty field. **Fig. 12** presents the uncertainty fields predicted by the BNN trained with varying amounts of data per experiment. All fields corresponded to the 12mm-cracked specimen at maximum displacement. As the number of training samples decreased, the overall magnitude of the epistemic uncertainty increased, reflecting greater epistemic uncertainty due to data scarcity. While the general spatial pattern of the epistemic uncertainty field remained consistent, its magnitude scaled with the amount of available training data. To clearly compare the overall magnitude of the epistemic uncertainty fields, the histograms and median values for four cases are shown in **Fig. 12**. As the number of training samples per experiment increased, the overall data distribution including median value shifted to the left, indicating a smaller magnitude of the overall epistemic uncertainty field. This trend supported the interpretation of epistemic uncertainty as an indicator of data sufficiency and model confidence. Taken together, these results indicated that the aleatoric and epistemic uncertainties quantified by the proposed framework each adhered to their theoretical definitions, demonstrating the validity of the two uncertainty fields.

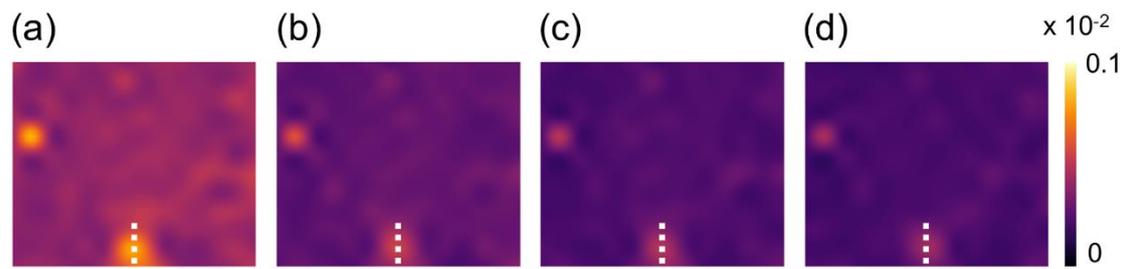

**Figure 12** Epistemic uncertainty fields under maximum displacement loading according to the number of training data used per experiment. White dashed lines represent the 12mm crack of the specimen. (a) 300; (b) 500; (c) 700; (d) 900.

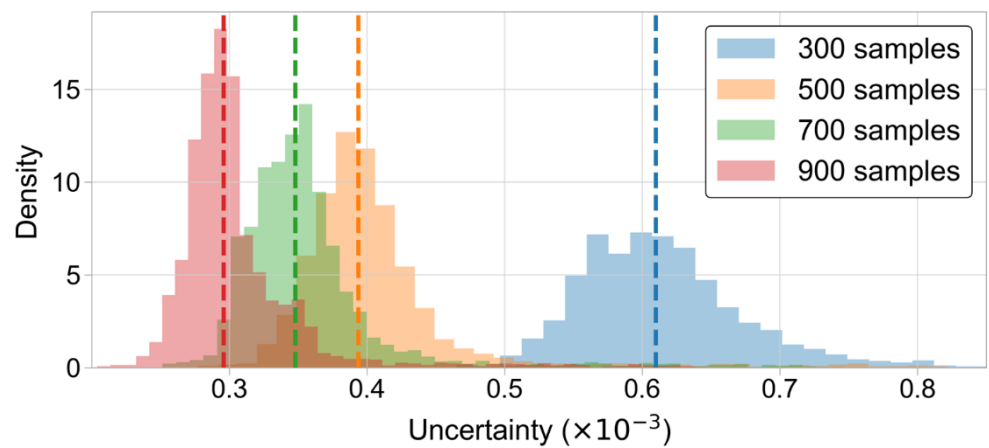

**Figure 13** Histograms and median values of epistemic uncertainty fields for four cases, each based on varying amounts of training data used per experiment.

# 5. Conclusion

This study presented a framework for real-time SHM based on a BNN. By applying PCA for dimensionality reduction, the proposed framework enabled efficient reconstruction of the full-field strain from sparse sensor measurements. The BNN was trained through a two-stage process consisting of pre-training and HMC sampling, enabling separate quantification of aleatoric and epistemic uncertainties and real-time estimation of full-field uncertainty fields. This explicit aleatoric-epistemic uncertainty representation enhanced both the interpretability and reliability of the reconstructed strain fields, providing insights into variability inherent to the data and into model-related confidence. The findings highlighted the potential of this approach to support more reliable and risk-aware DT systems for SHM.

The main findings of this study are as follows:

(1) The BNN framework combined with PCA successfully reconstructed the full-field strain from sparse strain gauge inputs and captured variations in the strain distribution corresponding to different crack lengths.

(2) Mode-wise predicted variances estimated during pre-training, were used directly as the likelihood variances in HMC. This choice reflected the relative importance of each mode and facilitated stable posterior sampling.

(3) The aleatoric uncertainty field derived during the pre-training process exhibited elevated levels near the crack tip and along the boundary regions. These high values indicated intrinsic limits of prediction imposed by measurement variability and crack-induced singularities.

(4) The epistemic uncertainty field obtained via HMC sampling reflected the time-varying risk of prediction failure due to a lack of model confidence and served as an indicator of data scarcity. Together with the aleatoric uncertainty field, it enabled distinction between data-inherent and model-related issues.

# CRediT authorship contribution statement

**Hanbin Cho:** Writing – review & editing, Writing – original draft, Software, Methodology, Investigation, Formal analysis, Data curation, Conceptualization. **Jechoen Yu:** Writing – review & editing, Methodology, Investigation. **Hyeonbin Moon:** Writing – review & editing, Methodology, Investigation. **Jiyoung Yoon:** Data curation. **Junheyong Lee:** Data curation, Methodology. **Giyoung Kim:** Data curation. **Jinhyoung Park:** Data curation. **Seunghwa Ryu**: Writing – review & editing, Validation, Supervision, Project administration, Funding acquisition, Conceptualization.

# Declaration of competing interest

The authors declare that they have no known competing financial interests or personal relationships that could have appeared to influence the work reported in this paper.

# Acknowledgements

This work was supported by the National Research Foundation of Korea (NRF) (No. RS-2025-16070951) and the InnoCORE program (N10250154), funded by the Ministry of Science and ICT. This research was also supported by a grant (RS-2023-00215667) from the Ministry of Food and Drug Safety.

# Data availability

Data will be made available on request.